\documentclass[conference]{IEEEtran}
\IEEEoverridecommandlockouts
\usepackage{cite}
\usepackage{amsmath,amssymb,amsfonts}
\usepackage{graphicx}
\usepackage{textcomp}
\usepackage{algorithm}
\usepackage{algpseudocode}
\usepackage{url}
\usepackage{svg}
\usepackage{epstopdf}
\usepackage[misc]{ifsym}
\usepackage{xcolor}
\usepackage{subcaption}
\usepackage{diagbox}
\usepackage{multirow}
\usepackage{tikz} 
\usetikzlibrary{shapes,arrows,positioning}
\usepackage{gensymb}
\usepackage{caption}
\usepackage{hyperref}
\algnewcommand\algorithmicintialize{\textbf{Initialize: }}
\algnewcommand\Initialize{\State\algorithmicintialize}

\algnewcommand\algorithmiccompute{\textbf{Compute: }}
\algnewcommand\Compute{\State\algorithmiccompute}

\def\BibTeX{{\rm B\kern-.05em{\sc i\kern-.025em b}\kern-.08em
    T\kern-.1667em\lower.7ex\hbox{E}\kern-.125emX}}
\begin{document}

\title{{Robust Proximity Operations using Probabilistic Markov Models}
\thanks{This work is supported by the Air Force Office of Scientific Research (AFOSR), as a part of the
SURI on OSAM project “Breaking the Launch Once Use Once Paradigm” (Grant No: FA9550-
22-1-0093).}
}

\author{Deep Parikh,\thanks{Authors are with Department of Aerospace Engineering, Texas A\&M University, College Station, TX. \{deep, alikhowaja\}@tamu.edu}
 Ali Hasnain Khowaja, Manoranjan Majji}

\newcommand{\deep}[1]{{\color{green}Deep: {#1}}}
\newcommand{\ali}[1]{{\color{orange}Ali: {#1}}}

\maketitle

\begin{abstract}
A Markov decision process-based state switching is devised, implemented, and analyzed for proximity operations of various autonomous vehicles. The framework contains a pose estimator along with a multi-state guidance algorithm. The unified pose estimator leverages the extended Kalman filter for the fusion of measurements from rate gyroscopes, monocular vision, and ultra-wideband radar sensors. It is also equipped with Mahalonobis distance-based outlier rejection and under-weighting of measurements for robust performance. The use of probabilistic Markov models to transition between various guidance modes is proposed to enable robust and efficient proximity operations. Finally, the framework is validated through an experimental analysis of the docking of two small satellites and the precision landing of an aerial vehicle. 
\end{abstract}
\begin{IEEEkeywords}
pose estimation, proximity operations, small satellites, aerial vehicles, Markov decision process
\end{IEEEkeywords}

\section{Introduction}
The recent technological advancements have enabled new avenues for utilizing various autonomous systems to tackle some of the prevelent technologiical challenges. For example, Unmanned Aerial Vehicles (UAV) in logistics/last-mile delivery\cite{su14010360,9869811}, and crewed aircraft for urban air mobility\cite{STRAUBINGER2020101852}. However, all such systems need some auxiliary sensors to determine the accurate position when nearing the intended drop-off/landing location for autonomous operations\cite{miranda2022autonomous}. This often involves switching from less accurate GPS-based localization to more accurate position estimates derived from auxiliary sensors like Ultra Wide-Band (UWB) radar and camera. In the field of robotics, there has been a pivotal shift in research and development of dexterous manipulation \& humanoids\cite{10415857}. For most of the applications involving grasping, there is often a need to switch from more coarse primary actuators to finer dexterous manipulators while nearing the object of interest\cite{9134855}.

On-orbit satellite servicing has been extensively studied for refurbishing and refueling satellites, construction of large structures in space, and orbital debris management \cite{osam_godd,king2001space}. Furthermore, the fast-paced advancements in the CubeSat technology have carved new frontiers in low-cost, low-risk space missions\cite{GAMBLE2014226}. More recently, some of the complex missions involving satellite swarming\cite{doi:10.2514/1.A35598} and rendezvous\cite{ROSCOE2018410} have also seen the involvement of CubeSats. However, the form factor, limited on-board power generation, and constrained fuel storage for CubeSats have proven to be a significant obstacle in achieving reliable and robust autonomy during proximity operations and docking applications\cite{spiegel2023cubesat}. 

\begin{figure}[!t]
\centerline{\includegraphics[width=0.4\textwidth]{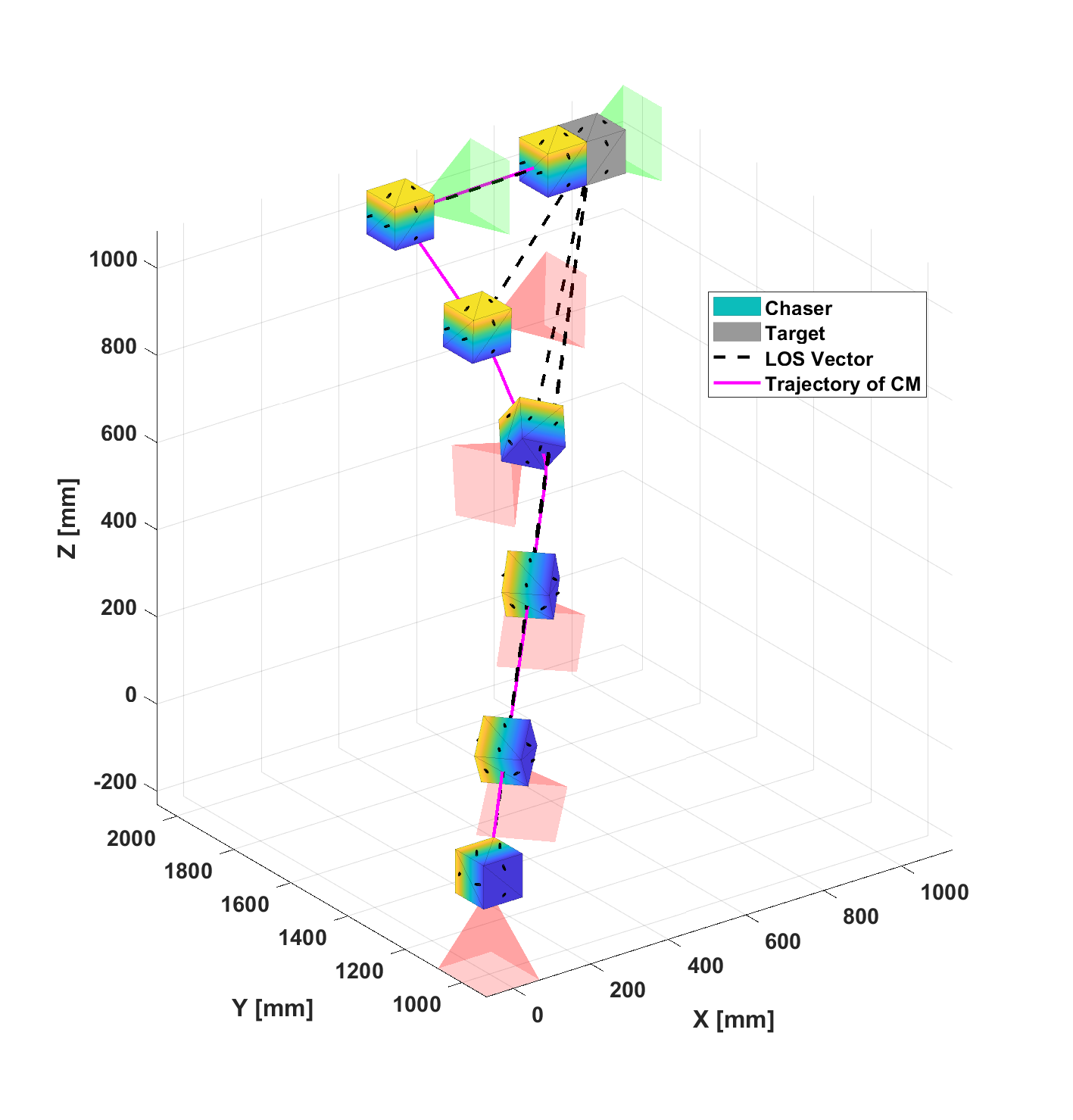}}
\caption{Three stage guidance algorithm for TPODS docking : The initial motion is in the direction of the target, followed by a reorientation to the desired docking attitude. Finally, the chaser aligns directly in front of the docking face and completes the docking. Availability of vision measurements is depicted with a change in the color of the camera FOV cone.}
\label{fig:guidance_TPODS}
\end{figure}

Almost all of such switching operations for sensors and actuators are performed with conservative bounds to ensure robustness against stochastic disturbances. This results in undesired consequences such as increased hovering time of an aerial vehicle and reduced practical operational range, along with high noise exposure to nearby inhabitants. Some of the previous studies have presented a covariance-based switching algorithm to handle transitions from high uncertainty to sudden availability of accurate measurements\cite{crassidis2009decentralized}. While such approaches offer an efficient avenue to obtain accurate pose estimates, they lack any insights on decisions that can drive position covariance. The intermediate decisions to sacrifice efficiency for better accuracy can be based on the current objectives and associated state uncertainties. Markov decision process(MDP) can be leveraged in such situations when a trade-off between efficiency and robustness is warranted\cite{bertuccelli2008robust}.

This paper presents a MDP based state machine to conduct robust proximity operations. Section \ref{sys_d} discusses the system dynamics with position-attitude coupling. The following section presents the sensor models for the IMU, UWB radar, and camera. The paper further introduces a unified pose estimator in Section \ref{sec_pose} followed by the guidance algorithm for proximity operations. The main contribution of this study is presented in Section \ref{sec_main} where a probabilistic Markov model based state machine is devised, implemented and analyzed for effectiveness, robustness and efficiency. Finally, Section \ref{sec_exp} includes exprimental validation of the proposed framework for two distinct applications : a) docking of two CubeSat motion emulators and b) precision landing of a UAV.

\begin{figure}[!b]
\centerline{\includegraphics[width=0.4\textwidth]{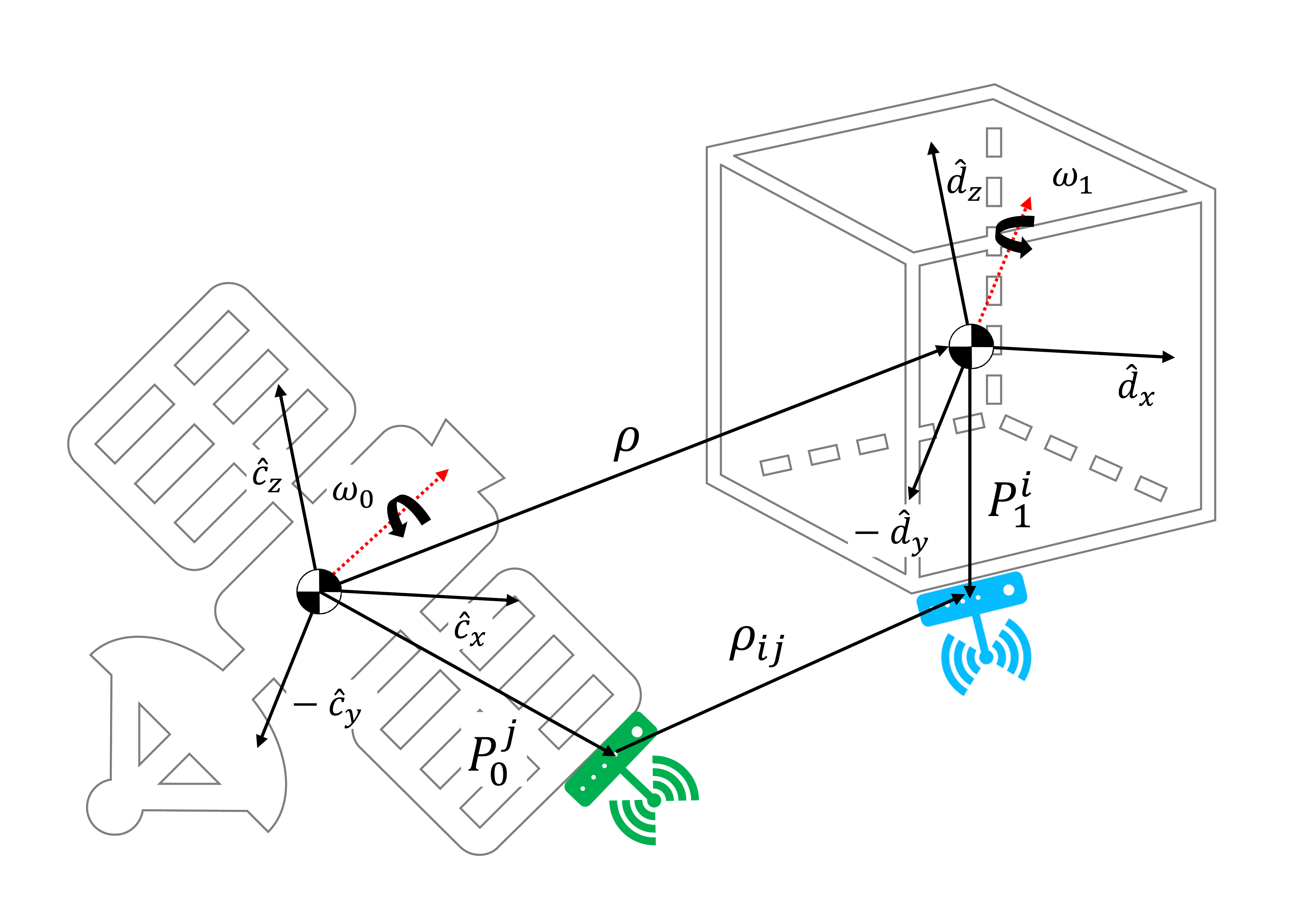}}
\caption{TPODS module uses UWB radar in two-way ranging mode to measure the distance to stationary anchors. Since the anchors and UWB sensor are not mounted at respective centers of mass, rotational and translation motion of the UWB sensor relative to the stationary anchors are coupled.}
\label{fig:TPODS_coupling}
\end{figure}

\section{Introduction and System Dynamics\label{sys_d}}
The Land, Air and Space Robotics(LASR) laboratory at Texas A\&M University is harnessing CubeSat technology to  gain control of RSO\cite{TPODS_detumble}. The critical technology advancement is embodied in the Transforming Proximity Operations and Docking System(TPODS) modules\cite{TPODS_system,TPODS_estm}. One of the crucial phases of a detumbling maneuver using TPODS is the proximity operation of modules to relocate them to the preferred location on the RSO after their deployment. \cite{TPODS_detumble}. 

For the analysis presented in this paper, a scenario consisting of a couple of CubeSat agents and a relatively stationary mother-ship satellite is considered\cite{TPODS_GNC24}. The objective is to make the chaser TPODS dock to a specified face of the stationary target TPODS. As shown in Fig.~\ref{fig:TPODS_coupling}, the mother-ship is equipped with multiple UWB transceivers. Each TPODS agent is also equipped with a UWB transceiver as well as a gyro and monocular camera. Since the UWB radar range sensors measure Time Of flight(TOF) \cite{OG_fusion}, the measurements are available for the relative distance $\boldsymbol{\rho_{ij}}$. Hence, the pose estimation algorithm needs to accurately predict the motion of the vector $\boldsymbol{\rho_{ij}}$ in the reference frame $\boldsymbol{\hat{c}}$, affixed with the mother-ship. The translation and rotational motion governing equations are\cite{ALFRIEND2010227}
\begin{align}
\boldsymbol{\ddot{\rho}_{ij}} &= \boldsymbol{\ddot{\rho}}  + \boldsymbol{\dot{\omega}} \times \boldsymbol{P_1^i} + \boldsymbol{\omega} \times \left(\boldsymbol{\omega} \times \boldsymbol{P_1^i}\right)  \label{eqn:coupled_EOM} \\
\boldsymbol{I_0 \dot{\omega}} &= \boldsymbol{I_0 D I_1^{-1}} \left[ \boldsymbol{N_1-D^T \omega} \times\ \boldsymbol{I_1 D^T \omega} \right]   \label{eqn:coupled_EOM_att}
\end{align}
Where $\boldsymbol{D}$ is the rotation matrix that transforms vectors from TPODS reference frame $\boldsymbol{\hat{d}}$ to the mother-ship frame $\boldsymbol{\hat{c}}$, $\boldsymbol{\omega}$ is the rotational angular velocity of TPODS relative to the mother-ship, $\boldsymbol{I_0}$ and $\boldsymbol{I_1}$ are respective inertia tensors for the mother-ship and TPODS, and $\boldsymbol{N_1}$ is the external torque applied to the TPODS. The CubeSat module has ten nozzles arranged in 'X' configuration on each opposite face to enable 6-DOF\cite{TPODS_GNC24}.

\section{Sensor models \label{sen_mod}}
The primary sensors considered for this study are rate gyroscope, UWB radar, and monocular vision. The sensor models for rate gyroscope and UWB radar are discussed in \cite{OG_fusion} and adopted in this paper. To generate synthetic measurements from the true state, white noise with standard deviations of $2\ cm$ is infused for the range measurements\footnote{\url{https://www.qorvo.com/products/d/da007948}}. It is considered that UWB radio modules mounted on four corners of the mother-ship having relative positions, $(x,y,z) = (1,-0.2,0.5); (1,0.2,-0.5); (-1,0.2,-0.5); (-1,-0.2,0.5)$m respectively. Another radio module is considered at a diagonal of TPODS module having $\boldsymbol{P_1^i}=(0.05,0.05,0)$m.  

For the rate gyro measurements the white noise with matching characteristic of VectorNav{\textregistered} VN-100 is infused\footnote{\url{https://www.vectornav.com/docs/default-source/product-brief/vn-100-product-brief.pdf}}. For the monocular vision measurements, pin-hole projection model is considered and the subsequent pixel locations of each feature point are generated as simulated measurements\cite{Ali_GNC23}. Measurement noise with a standard deviation of $1\ mm$ is injected after converting the pixel locations of feature points to their respective position vectors in the camera frame and fed to the pose estimation algorithm \cite{Ali_GNC24}.

\section{Navigation and Control\label{sec_pose}}
A discrete extended Kalman filter(EKF) based estimator is leveraged to estimate the state of the TPODS module. The discrete EKF formulation of \cite{crassidis2011optimal} is adopted for this paper and the equations are not repeated here for the sake of brevity. The goal of the estimator is to combine the measurements from the rate gyroscope, UWB radar, and monocular vision along with the knowledge about system dynamics presented in Section \ref{sys_d}, sensor models of Section \ref{sen_mod}, and the current control input to compute the best guess of the current relative position, velocity and orientation of the TPODS module.

\subsection{Pose Estimator}
A caret sign represents an estimated value of each quantity. It is assumed that unbiased measurements for the angular velocity $\boldsymbol{\omega}$ are available with the respective measurement noise, and a quaternion $\boldsymbol{q}$ tracks the attitude of a TPODS module relative to the mother-ship. 
\begin{equation}
\boldsymbol{\hat{\omega}} = \boldsymbol{z}_{gyro}
\end{equation}

The estimator predicts a stochastic state vector,
\begin{equation}
\boldsymbol{\zeta} = (\boldsymbol{\rho_{ij}},\boldsymbol{\dot{\rho_{ij}}},\boldsymbol{\delta q})
\end{equation}
Where $\boldsymbol{\delta q}$ is a three-parameter vector representing errors in the estimated attitude. The selected structure of the pose estimator offers two distinct advantages
\begin{table}[t]
    \centering
    \caption{Computational performance of a single prediction and measurement update step of the pose estimator with single range and angular velocity measurements. Generated on a computer with $11^{th}$ Gen Intel\textregistered Core\texttrademark i5-1135G7 @ 2.40GHz, 16 GB RAM, for running a MATLAB\textregistered script.}
    \begin{tabular}{|c|c|c|c|}
    \hline
    \backslashbox{Structure}{Metric} & FLOP & Memory (kb) & Runtime (ms)\\\hline
     Compounded & 210,421 & 644  & 7.054\\\hline
     Tandem & 100,009 & 128  & 3.407\\\hline
    \end{tabular}   
    \label{tab:compute_compare}
\end{table}

\begin{enumerate}
    \item The estimator is divided into two tandem structure, one for tracking six transnational states and another for three attitude error parameters. As the computational cost of the extended Kalman filter is proportional to the cube of filter states\cite{filter_compute}, the tandem structure helps in lowering the total computational cost of the algorithm.  
    \item Instead of tracking the relative attitude via quaternion, the three attitude error parameters mitigate some of the adverse effects of quaternion normalization constraint step\cite{crassidis2011optimal}. The estimated attitude is recovered with a reference quaternion from the attitude error parameters with $\mathbf{\delta q} = \mathbf{q}\otimes\mathbf{\hat{q}}^{-1}$.
\end{enumerate}

\subsection{Attitude and Position Control}
To drive the TPODS module to the desired pose, an attitude and a position controller act in tandem. The control algorithm utilizes the estimated states, i.e, reference quaternion $\boldsymbol{\hat{q}}_{ref}$, relative position and velocity ($\boldsymbol{\hat{\rho},\dot{\hat{\rho}}}$) of the center of mass (obtained from the kinematic relationship between $\boldsymbol{\rho}$ and $\boldsymbol{\rho}_{ij}$). A closed-loop attitude controller acts on errors in desired and estimated quaternion\cite{wie1985quaternion}. Such control structure avoids any intermediate conversions from quaternion attitude representation to any other attitude parameters like Euler angles. This provides robustness against well-known pitfalls of singular rotation matrices. 

For the translation degrees of freedom, a full-state feedback discrete Linear Quadratic Regulator(LQR) with feed-forward and integral term is implemented\cite{ogata1995discrete}. The plant model defined by Equation~\eqref{eqn:coupled_EOM} is linearized around the stationary operating point and the respective steady-state gains are computed. The feed-forward terms provide additional excitation to drive the states to desired non-zero values and the integral terms provide robustness against any unmeasured disturbances\cite{dorato1992robust}.

\subsection{Truth propagation and State Prediction}
\begin{figure}[t]
\centerline{\includegraphics[width=0.45\textwidth]{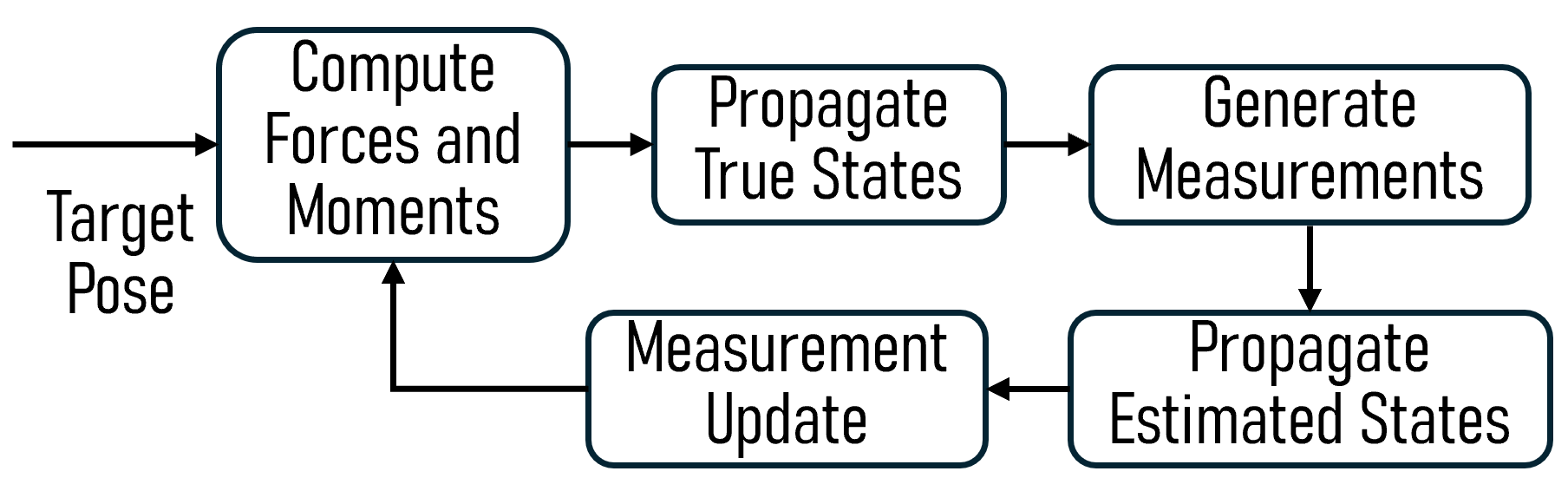}}
\caption{Simulation setup : The true states consist of relative position and velocity (expressed in reference frame $\boldsymbol{\hat{c}}$), attitude quaternion, and angular velocities (expressed in reference frame $\boldsymbol{\hat{d}}$). Generated measurements are relative range to a specific anchor, angular velocities, and location of feature points if the target is within camera FOV.}
\label{fig:flow_diag}
\end{figure}

For the evaluation of the pose estimator, a simulation setup is implemented as presented in Fig.~\ref{fig:flow_diag}. Starting with an initial guess of the mean of the states, the system dynamics given by Equations~\eqref{eqn:coupled_EOM} and ~\eqref{eqn:coupled_EOM_att} are utilized to predict the mean of states at the next time instance. Similarly, the system dynamics Jacobian is computed and used with the current state covariance and process noise to predict the state covariance for the next time instance\cite{crassidis2011optimal,TPODS_GNC24}.

\subsection{Measurement update and quaternion reset}
Since angular velocity is not an explicit state of the estimator, the rate measurements from the gyroscope do not have a direct impact during the measurement update. However, the state propagation requires knowledge of angular rates and current orientation. Once the attitude error vector is propagated, the quaternion reset step is executed to update the reference quaternion $\boldsymbol{\hat{q}}_{ref}$ and reset the attitude error vector back to zero\cite{Markley}. 

A distinctive characteristic of UWB radar measurements is the presence of outliers due to clock drift and multi-path reflection between two radar modules\cite{9372785}. The Mahalanobis distance provides a degree of statistical agreement of the measurements based on the known plant and sensor models\cite{DEMAESSCHALCK20001}. Hence, an outlier rejection scheme based on the Mahalanobis distance has been chosen to achieve robust state estimates. A squared Mahalanobis distance is computed for each measurement, and the individual measurements that fall outside the $99.99\%$ probability gate are rejected during the measurement update\cite{TPODS_GNC24}. 

The vision-based pose estimation employs sensor fusion of the camera with IMU measurements via an EKF. The camera sensor uses features, such as LED markers, for establishing 3D-2D correspondences\cite{Ali_GNC23,Ali_GNC24}. The three-dimensional coordinates of the markers are defined in the body-centered coordinate frame of the target TPODS module. Although, for perspective-n-point (PnP) projection-based pose computation a minimum of three points are required, three additional markers are used to form a pattern on each face of the TPODS module to ensure good observability at closer distances. 

Once within a specified range, the LOS vector between the chaser and the target is computed. The measurements for six feature points are generated if the vector lies within the Filed Of View(FOV) cone of the camera sensor. It is considered that each measurement vector has the same associated uncertainty and the location of each feature in the target reference frame is known with high accuracy.

\section{Guidance Algorithm and State Machine\label{sec_guidance}}
The docking of TPODS is accomplished by a three-stage guidance algorithm, visually shown in Fig.~\ref{fig:guidance_TPODS}. It is often beneficial to continue in a specified orientation due to certain mission objectives and constraints such as maximizing solar power generation, protecting sensors from direct sun exposure, or maintaining visual/radio coverage of certain terrestrial locations. However, such orientation might not be beneficial from the perspective of navigation. As depicted in Fig.~\ref{fig:guidance_TPODS}, the target might be outside camera FOV during the motion. Hence there is a strong motivation to switch orientations late in the motion, while still ensuring a reliable docking.

 The switching between multiple guidance modes is handled by a central state-machine which tracks the relative range between the chaser and target. However, since the estimated position of the chaser originates from stochastic measurements with non-zero uncertainties, it is not trivial to determine a robust and efficient set of switching distances that can ensure the docking objective is fulfilled for a wide range of estimation errors. Often, this necessitates the switching distances to be conservative, resulting in an increased total time and fuel/energy consumption. Furthermore, an increase in hovering time directly results to a drastic increase in energy consumption for aerial vehicles\cite{abdilla2015power}.

\begin{figure}[!t] 
\centering 
\begin{tikzpicture}[->,>=stealth',shorten >=2pt, line width=0.5pt, node distance=3cm]

\node [circle, draw] (zero) {\scriptsize LOS}; 
\path (zero) edge [loop above] node {\scriptsize $A_{11}$} (zero); 
\node [circle, draw] (one) [right of=zero] {\scriptsize Reorient}; 
\path (zero) edge node[above] {\scriptsize $A_{12}$} (one); 
\node [circle, draw] (two) [right of=one] {\scriptsize Align}; 
\path (one) edge [loop above] node {\scriptsize $A_{22}$} (one); 
\path (one) edge[bend left] node[above]{\scriptsize $A_{23}$} (two);
\path (two) edge [loop above] node {\scriptsize $A_{33}$} (two); 
\path (two) edge[bend left] node[below]{\scriptsize $A_{32}$} (one); 

\end{tikzpicture} 
\caption{State machine for docking guidance along with respective transition probabilities. Since each $A_{ij}$ represents a probability, they have to be non-dimensional and their magnitude must lie between 0 and 1. In addition, the total probability of all outgoing arrows from a state must add to 1.} 
\label{fig:state_machine} 
\end{figure}
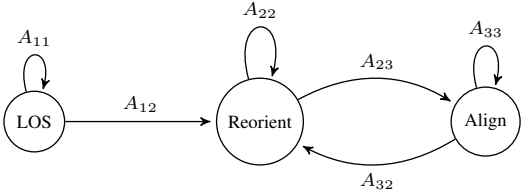

\subsection{Markov Decision Process\label{MDP}}
Instead of having predefined switching distances, a dynamic assignment of the switching distances can enable a pragmatic and effective trade-off between optimality/mission objectives and robustness. Some of the earlier explorations in this have based their approach on the covariance of the estimator states\cite{crassidis2009decentralized}. However, the scenario considered in this paper also has additional decision variables such as relative range to target, orientation misalignment, and the possibility of improving accuracy of pose estimates by aligning to the target. Hence, instead of just relying on the covariance for decision-making, MDP can be employed to dynamically assign the switching distances to achieve multiple performance matrices \cite{norris1998markov,bertuccelli2008robust}.

\subsection{Probability Assignment Rules\label{sec_main}}
For the effective switching of the state machine presented in Fig.~\ref{fig:state_machine}, the transition probabilities are a key factor. The constraints on $A_{ij}$, representing a probability, necessitate intricate value assignment rules at each instance of decision-making. The overall scheme developed to assign these probabilities is presented in Algorithm~\ref{alg:state_switch}. The operation begins with aggressive values of switching distances $r_1$ and $r_2$. As mentioned in the previous section, the first phase of the motion is in the general direction of the target LOS vector. Once the chaser crosses the predefined threshold distance $r_t$, the MDP-based switching algorithm is activated.

\begin{algorithm}[b]
\caption{Transition Probabilities and Switching}\label{alg:state_switch}
\begin{algorithmic}
\Initialize $r_1 \gets 0.1,\ r_2 \gets 0.3$ 
\While{$r \leq r_t \ \&\&\  r > r_1 $}
\If{$r > r_2$}
    \State $A_{11} \gets\left(\frac{r}{r_t}\right)*\left(\frac{1}{1+d}\right)*\left(\frac{1}{\left\lVert\left[q_1\ q_2\ q_3\right]_e\right\rVert}\right) $ 
    \If{$A_{11} \leq 0.4$}
    \State $r_2 \gets r$
    \EndIf
\Else
    \State $A_{22} \gets \left(\frac{r}{r_2}\right)*\left(\frac{1}{1+d}\right) \quad A_{23}\gets 1-A_{22}$
    \State $A_{32} \gets \left(\frac{max\left(\sigma_{pos}\right)}{\sigma_{UWB}}\right)*\left(\frac{r_2-r}{r_2-r_d}\right) \quad A_{33}\gets 1-A_{32}$
    \Compute Steady state probabilities $\left[\pi_1 \ \ \pi_2\right] $for \\
    \hspace{3cm}$A_n = \left[\begin{smallmatrix}
            A_{22} & A_{23} \\
            A_{32} & A_{33} \\
            \end{smallmatrix}\right]$
    \If{$\pi_{2} \leq 0.6$}
    \State $r_1 \gets r$
    \EndIf
\EndIf
\EndWhile
\end{algorithmic}
\end{algorithm}

The chaser is allowed to keep pressing forward in the general direction of the LOS vector till the filter
\begin{itemize}
    \item Is producing consistent estimates $\to$ inferred from Mahalonobis distance $d$ 
    \item Has a sufficient distance to the target left to reorient $\to$ inferred from the normalized distance factor $r/r_t$, and the norm of vector component of quaternion error between the target orientation and chaser orientation $\lVert\left[q_1\ q_2\ q_3\right]_e\rVert$.
\end{itemize}
Due to the normalization of distance, and use of non-dimensionalized parameters, the switching factor $A_{11}$ remains dimensionless and close to 1. If the factor goes below a predefined threshold due to inconsistent and less accurate pose estimates, or less distance to target for reorientation, the guidance mode is switched to reorientation by setting the current range to the target as $r_2$.

In the second guidance mode, the chaser continues the motion in the direction of the LOS vector, albeit now oriented similarly to the target. The switching algorithm now checks for the trade-off between continuing the current motion and aligning in-front of the target, allowing more accurate vision measurement to improve pose estimates. However, as discussed earlier, retaining the current direction of motion might be beneficial for certain mission objectives. This is achieved by monitoring the maximum standard deviation in position estimates $\sigma_{pos}$ as a fraction of the standard deviation of the UWB sensor $\sigma_{UWB}$. For the instances when vision measurements are available early, the factor $A_{32}$ remains low as $\sigma_{pos}$ with vision measurements is significantly lower than $\sigma_{UWB}$, and distance traveled from $r_2$ is significantly lower than the total distance to be traveled from $r_2$ till docking distance $r_d$. 

However, when the vision measurements are not available due to FOV misalignment near the target, $A_{32}$ increases, driving the steady state probability of the state `\textit{Align}' low. This indicates a low confidence of the state machine to successfully align the chaser in-front of the target if the motion is continued in the same direction. Hence, the current radial distance is picked as the alignment radius. The alignment point is calculated, which is directly in-front of the docking face in the target reference frame. Finally, the chaser is commanded to move to the alignment point to acquire vision measurements.

\subsection{Monte Carlo Simulations}
\begin{figure}[t!]
     \begin{subfigure}[b]{0.24\textwidth}
        \centering
         \includegraphics[width=\textwidth]{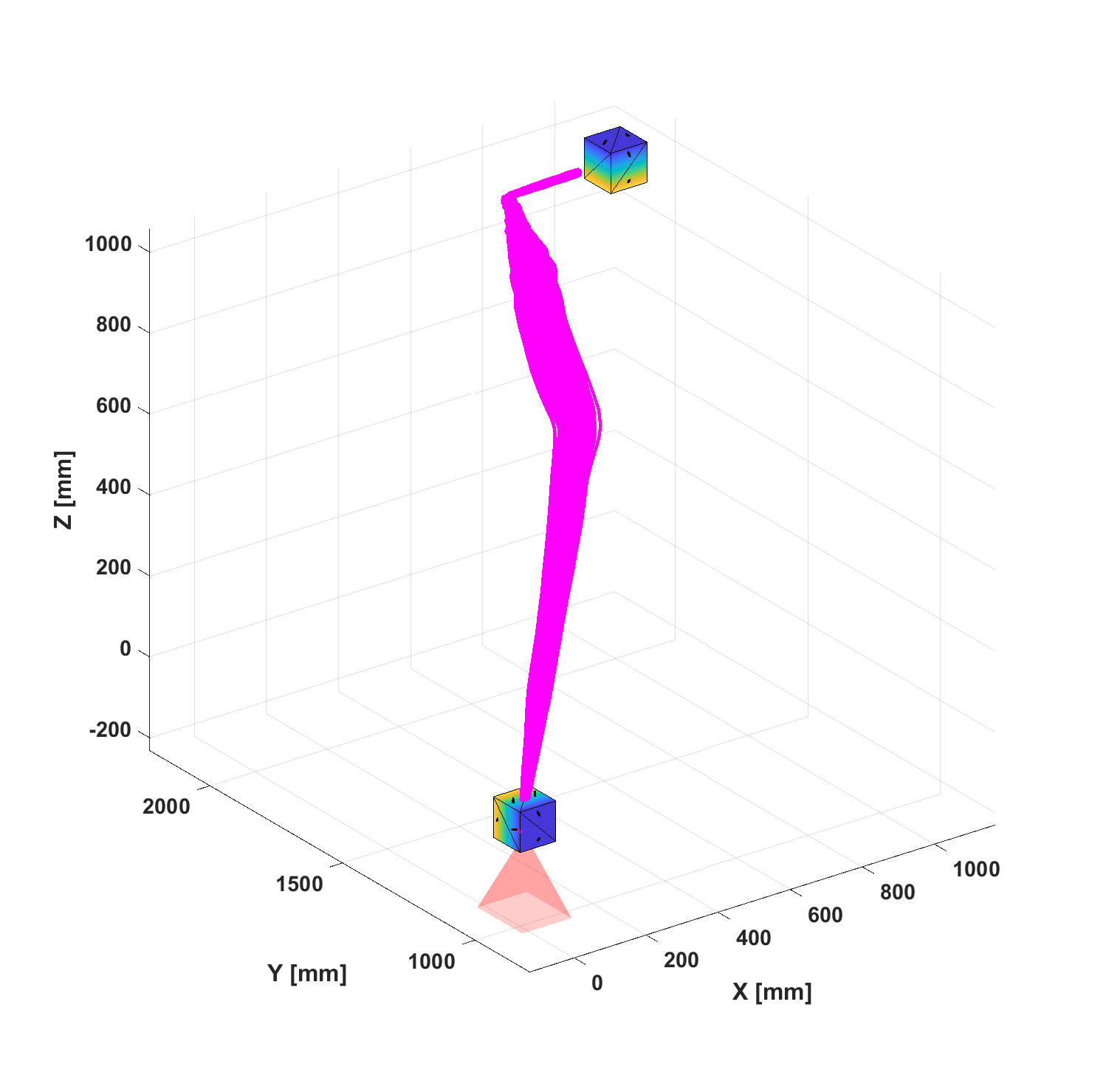}
         \caption{Fixed Switching}
     \end{subfigure}
     \centering
     \begin{subfigure}[b]{0.24\textwidth}
        \centering
         \includegraphics[width=\textwidth]{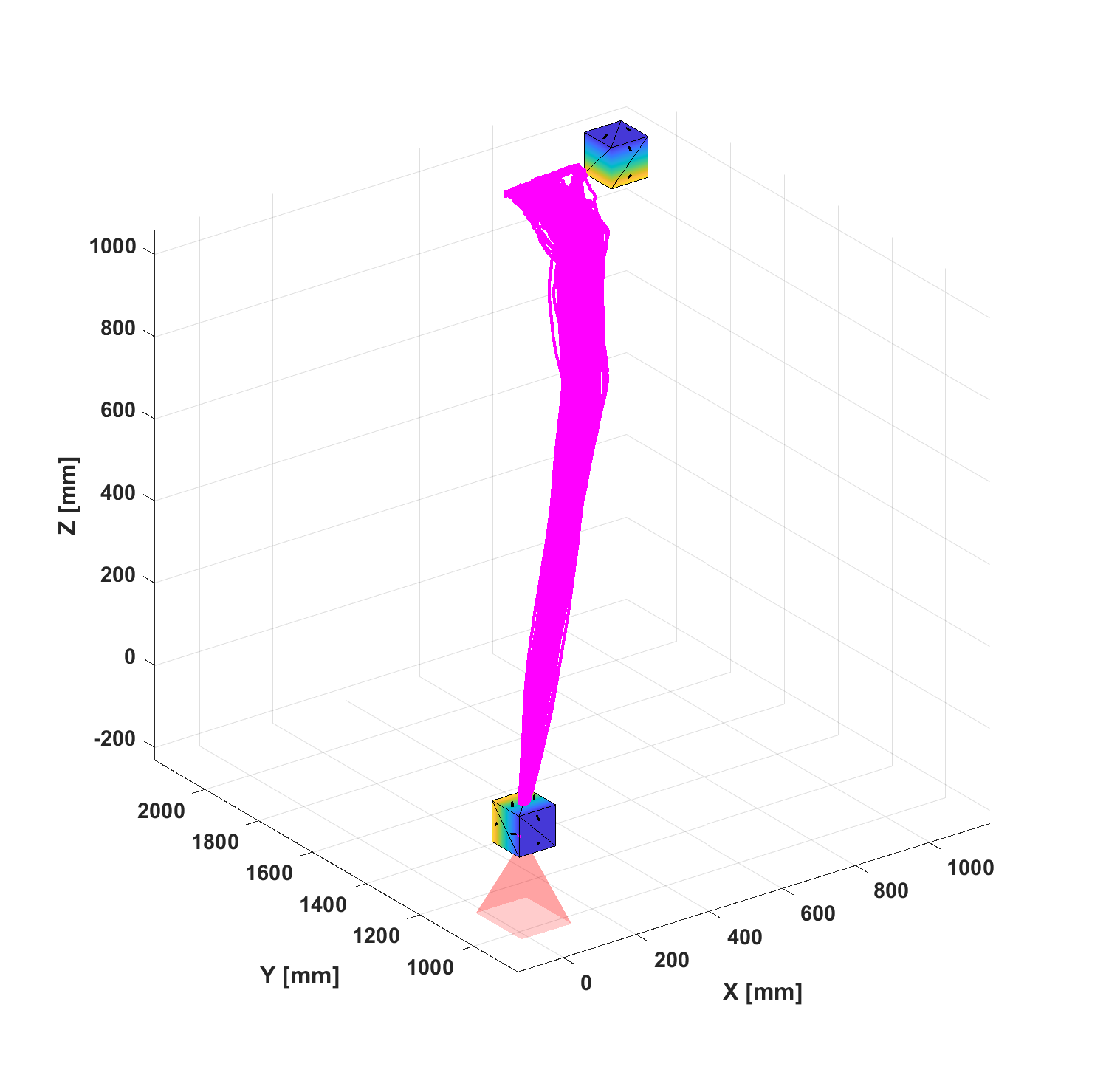}
         \caption{Adaptive Switching}
    \end{subfigure}
    \caption{Comparison of trajectories with fixed and adaptive switching distances for 500 run Monte-Carlo simulation. The target is located at $(1,2,1)m$ having the orientation $\left(\theta_x,\theta_y,\theta_z\right)=\left(-90\degree,0\degree,-90\degree\right)$, while the chaser is located at $(0,1,0)m$ with the orientation $\left(0\degree,0\degree,0\degree\right)$. The camera sensor is mounted at the bottom face of the chaser.}
    \label{fig:MC_traj}
\end{figure}

In order to assess the effectiveness of the proposed adaptive switching, extensive simulations have been carried out. The effect of adaptive guidance mode switching is evident in Fig.~\ref{fig:MC_traj}. The trajectories for fixed switching distances of $r_2$=$0.8$ \& $r_1$=$0.3$ show a conservative envelope as each instance is treated equally. The Majority of the dispersion appears during the reorientation phase and the motion becomes fairly deterministic near the alignment point. However, in the case of adaptive switching, the decision to reorient is taken deep into the motion, followed by alignment. Though this has an effect of carrying the dispersion deep into the motion, each instance is weighted based on the probability of successful docking.

\begin{figure}[!b]
\centerline{\includegraphics[width=0.49\textwidth]{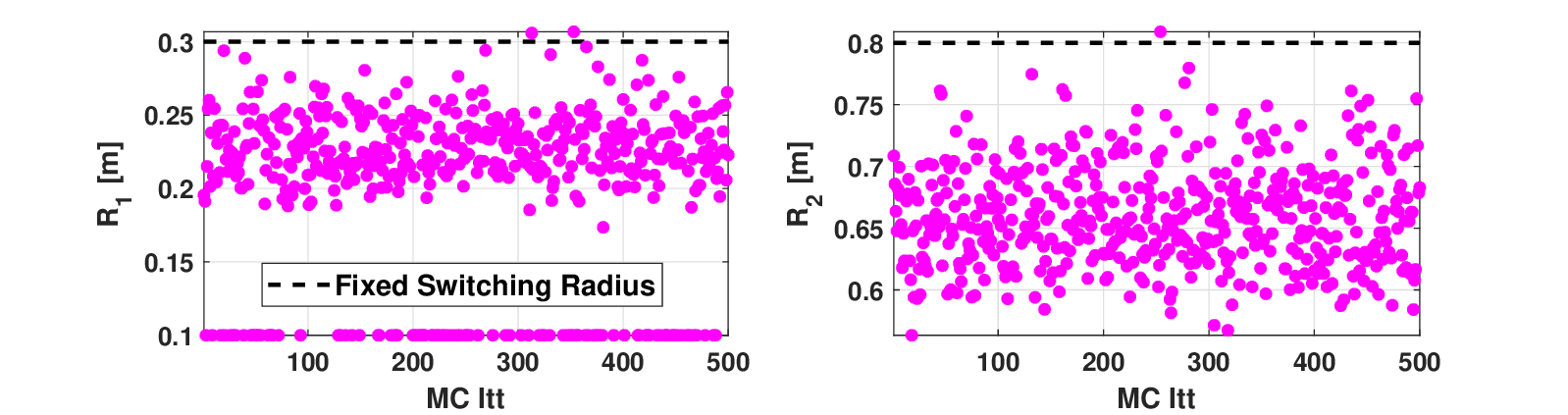}}
\caption{Switching distances picked by the proposed algorithm for 500 run Monte-Carlo simulations of the docking scenario of Fig.~\ref{fig:MC_traj}}
\label{fig:MC_swt_dis}
\end{figure}
A few performance metrics are presented in Table~\ref{tab:performance_MC} and Fig.~\ref{fig:MC_swt_dis}. The spread of switching distances in Fig.~\ref{fig:MC_swt_dis} along with the average combined errors at docking, presented in the first and second column of Table~\ref{tab:performance_MC}, underscores the effectiveness of the state machine in picking the switching distances ensuring successful docking. The Conservative switching distances of $r_2$=$0.8$ \& $r_1$=$0.3$ are inferred from Fig.~\ref{fig:MC_swt_dis} as the worst-case switching distances picked by the algorithm. Finally, the main contribution of the algorithm is evident when comparing the average docking time and average total impulse imparted during the operation. On average, a $10\%$ reduction in docking time and a $14\%$ reduction in total impulse is achieved with adaptive guidance mode switching.

\begin{table}[t!]
    \caption{Performance metrics for 500 run Monte-Carlo simulations of the docking scenario of Fig.~\ref{fig:MC_traj}}
    \setlength{\tabcolsep}{1pt}
    \centering
    \begin{tabular}{|c|m{0.14\linewidth}|m{0.14\linewidth}|m{0.12\linewidth}|m{0.18\linewidth}|}
    \hline
    \backslashbox{Switching}{Metric} & Position Error [cm] & Attitude Error [rad] & Docking Time [s] & Total Impulse [Ns]\\\hline
     Fixed distance & \centering 0.567 & \centering 0.0137 & \centering 140.05 & \quad 10.7858 \\\hline
     MDP & \centering 0.568 & \centering 0.0129 & \centering  126.65 & \quad \ 9.2679\\\hline
    \end{tabular}
    \label{tab:performance_MC}
\end{table}

\subsection{Steady-State Probabilities for Optimal Module Deployment}
Some of the earlier studies have provided insights into the impact of the general placement of UWB anchors on the overall accuracy of the position estimates \cite{5399478,7353810}. Based on the placement of the UWB anchors, best and worst-case steady-state position covariance can be determined. This information can be further utilized to find steady-state probabilities of various stages of the state machine using probability assignment rules of Algorithm~\ref{alg:state_switch}. 

The knowledge of steady state probabilities of the state machine provides significant insights into the chances of achieving mission objectives for a given set of system parameters like the placement of UWB anchors or initial pose relative to the target. Such information becomes vital in maximizing the overall success rate of achieving the end objective under the worst possible measurement noise scenarios. While earlier discussions have been focused on general fuel/energy optimality of de-tumbling an RSO using CubeSats modules deployed from different relative locations\cite{TPODS_detumble}, the framework presented in this paper enables novel avenues to analyze various scenarios for optimality in the sense of navigation. 

\section{Experimental Validation\label{sec_exp}}
The proposed pose estimation, guidance, and control framework is experimentally validated using a couple of scenarios involving proximity operations. In both applications, a vehicle is equipped with a primary long-range but less accurate sensor and a highly accurate secondary but short-range sensor. The algorithm is tasked to determine the switching parameters to ensure robustness against stochastic disturbances in meeting the overall objective.

\subsection{Docking of TPODS Modules}
The TPODS module MK-E\cite{TPODS_system,TPODS_estm,TPODS_GNC24}, which has been devised and fabricated at the LASR laboratory, is used as a chaser spacecraft and a stationary target. Four Loco Positioning nodes are used as anchors and the module is also equipped with a similar positioning node. The camera is commanded to initialize the pose estimation once in close proximity to the target. For long-range operations, only UWB radio and IMU measurements are used. 

\begin{figure}[!t]
\centerline{\includegraphics[width=0.4\textwidth]{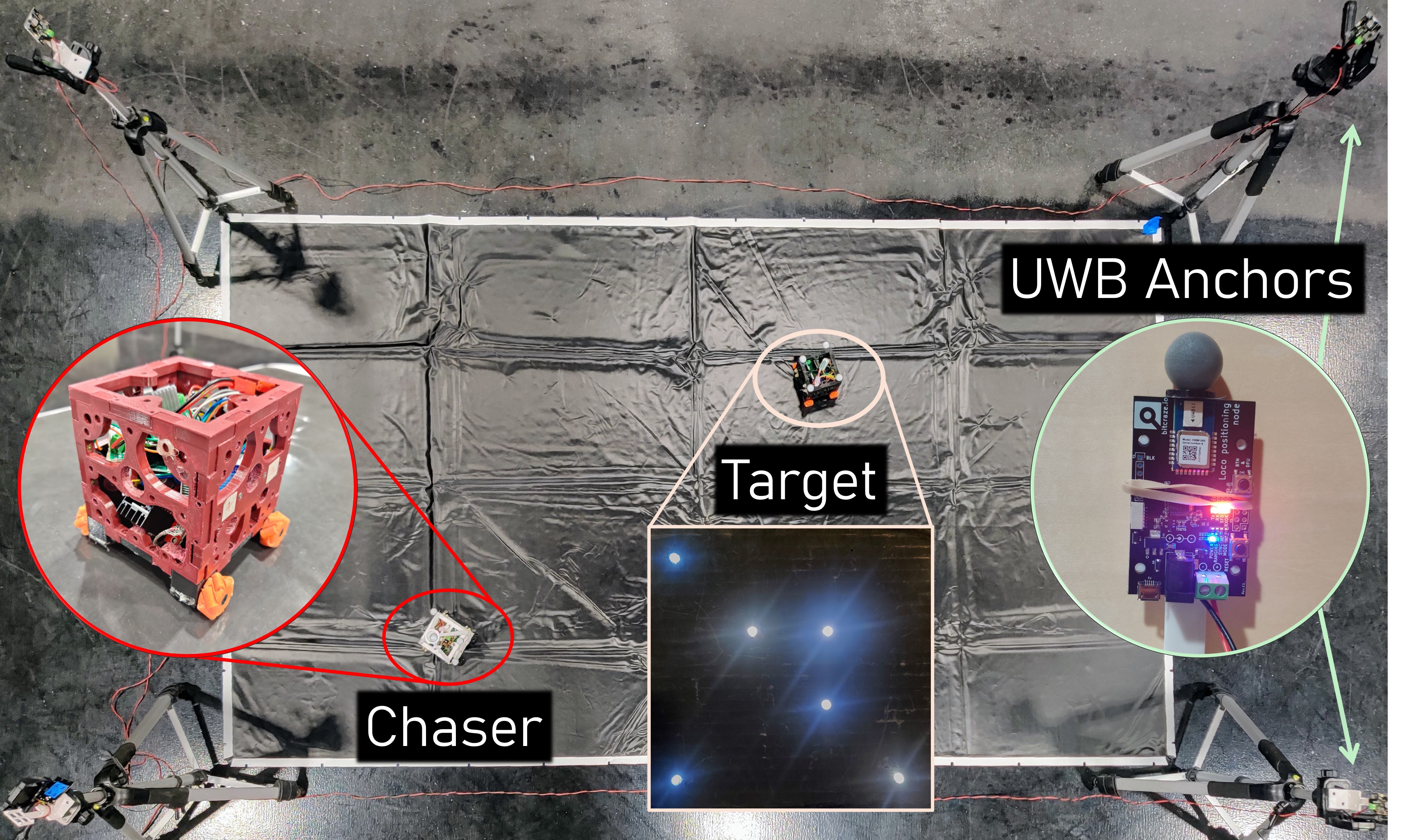}}
\caption{The MK-E version of TPODS consists of four N-20 electric drives along with a 3-D printed mecanum wheels, designed and printed as a single piece in-house. The module also consists VectorNav® VN-100 IMU sensor, Loco positioning node and OpenMv H7 monocular vision camera. Teensy 4.1 serves as the central computing unit for the TPODS module.}
\label{fig:Exp_setup}
\end{figure}

\begin{figure}[!b]
\centerline{\includegraphics[width=0.4\textwidth]{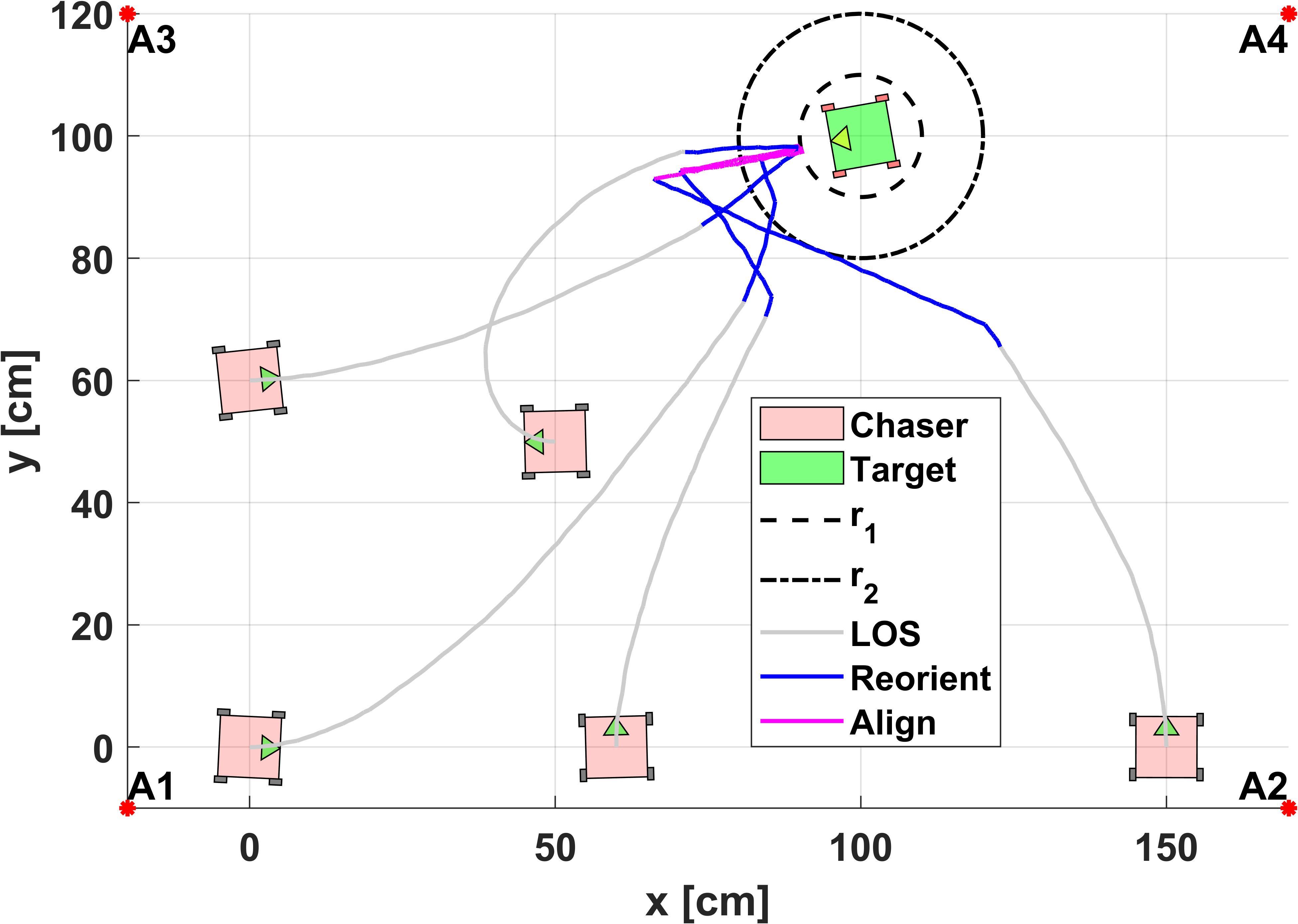}}
\caption{The chaser module is placed at several initial positions within the convex hull of four UWB anchors. The chaser begins motion towards the stationary target once the initial position is estimated with specified variance. On-board pose estimator and guidance algorithm drive the chaser towards the stationary target, while switching between different guidance modes.}
\label{fig:Exp_TPODS_res}
\end{figure}

 The computer requests orientation and angular rate from the IMU at $100Hz$, UWB range data at approximately $50Hz$, and the monocular vision-based pose estimate at $20Hz$. The estimator fuses the UWB range with the IMU data to estimate the pose of the module for long-range operations. Once within the vicinity of the target, the camera is commanded to initialize pose estimation and treated as the primary measurement along with IMU measurements. The pose from motion capture system is logged at $120Hz$ and further used as a ground truth for the performance evaluation.

The effectiveness of the algorithm is evident from the results presented in Fig.~\ref{fig:Exp_TPODS_res}. The on-board algorithm picks aggressive switching distance when the vision measurements made available early due to the alignment of the chaser and the target, eventually skipping the align guidance mode altogether. For other scenarios, switching distance are selected by the algorithm to ensure sufficient time for the chaser to align with the target face and complete the docking maneuver. 

\subsection{Precision Landing of an Aerial Vehicle}

\begin{figure}[t!]
\centerline{\includegraphics[width=0.4\textwidth]{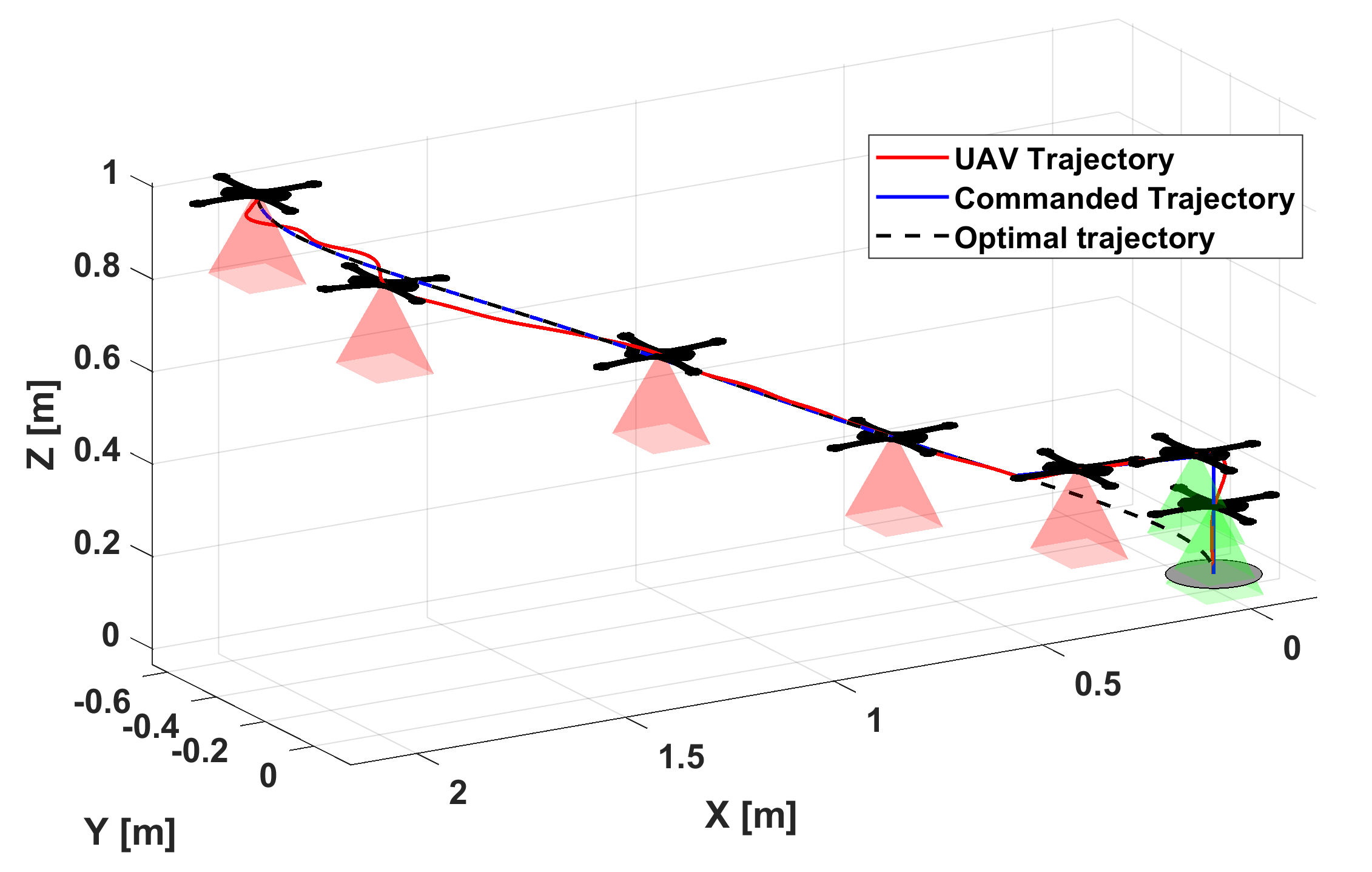}}
\caption{The UAV follows an optimal trajectory for the majority of the motion. The algorithm determines a deviation point based on the quality of pose estimates, availability of the vision measurements, and remaining distance from the pad.}
\label{fig:quad_single}
\end{figure}

A similar experimental analysis has also been conducted for the precision landing of quadrotor aerial vehicles. The objective is to follow energy-optimal landing trajectories for the majority of the motion and only deviate sufficiently to ensure a precise landing. The initial motion of the vehicle is driven by less accurate pose estimates, obtained from IMU odometry. Vision measurements are available for the final phase of the motion, resulting in precise pose estimates. As shown in Figure~\ref{fig:quad_single}, the energy optimal landing trajectories for quadrotor vehicles contain a large lateral movement followed by a brief vertical motion near the landing zone\cite{7487285}. However, such an approach is not preferred if the objective is to land precisely, driven by the availability of vision measurements. Consequently, the vehicle needs to be diverted to a suitable point which ensures availability of vision measurements. 

\begin{figure}[b!]
\centerline{\includegraphics[width=0.4\textwidth]{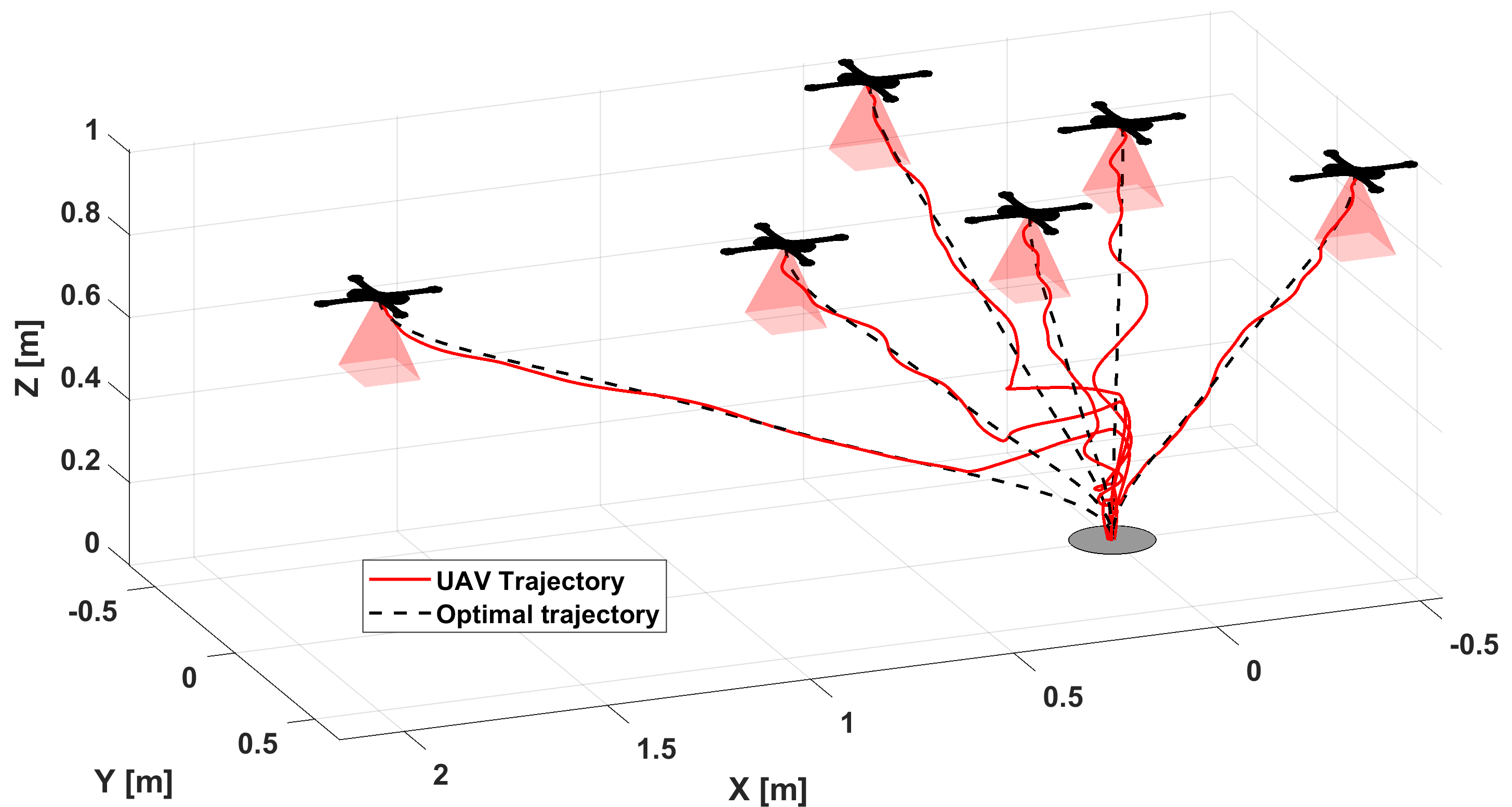}}
\caption{Effectiveness of the switching algorithm in ensuring precision landing.}
\label{fig:quad_many}
\end{figure}

\section{Conclusion}
A robust, low-cost pose estimation via sensor fusion along with adaptive switching of multi-state guidance modes have been proposed and validated in this paper. The limitation of unreliable pose estimates from primary long-range sensors while operating within proximity of the target is addressed by adding secondary short-range, but highly accurate measurements. A probabilistic Markov model to switch between various guidance modes is introduced along with effective rules to assign transition probabilities. Consequently, adaptation of the methods proposed in the paper are anticipated to provide a framework for robust and efficient proximity operations that requires a trade-off between extremizing different costs, while still ensuring the final objective is achieved.  

\section*{Acknowledgment}
Program monitors for the AFOSR SURI on OSAM, Dr. Andrew Sinclair  and Mr. Matthew Cleal of AFRL are gratefully acknowledged for their watchful guidance. Prof. Howie Choset of CMU, Mr. Andy Kwas of Northrop Grumman Space Systems and Prof. Rafael Fierro of UNM are acknowledged for their motivation, technical support, and discussions. 

\bibliographystyle{ieeetr}
\bibliography{./bib/refsm-astro}

\end{document}